\documentclass[runningheads]{llncs}
\usepackage{graphicx}
\usepackage{amsmath}
\begin{document}
\title{Interpreting Forecasted Vital Signs Using N-BEATS in Sepsis Patients}
%
%
\author{Anubhav Bhatti \inst{1}\orcidID{0000-0002-0035-9860} \and
Naveen Thangavelu \inst{2}\orcidID{0009-0005-9277-2909} \and
Marium Hassan \inst{1}\orcidID{0009-0004-4577-3362} \and
Choongmin Kim \inst{3}\orcidID{0000-0001-5812-889X} \and
San	Lee \inst{1}\orcidID{0009-0001-6488-7988} \and
Yonghwan Kim \inst{3}\orcidID{0000-0003-3022-6847} \and
Jang Yong Kim  \inst{4}\orcidID{0000-0001-8437-9254}
}
\authorrunning{A. Bhatti et al.}
%
\institute{ 
SpassMed Inc., Ontario, Canada \\
\email{\{anubhav.bhatti, marium.hassan, sanlee\}@spassmed.ca} \and
University of Toronto, Ontario, Canada \\
\email{naveen.thangavelu@mail.utoronto.ca} \and
Spass Inc., Seoul, South Korea\\
\email{\{cmkim, kyh\}@spass.ai} \and
St. Mary’s Hospital, Seoul, Korea \\
\email{vasculakim@catholic.ac.kr}
}
\maketitle              
\begin{abstract}
Detecting and predicting septic shock early is crucial for the best possible outcome for patients. Accurately forecasting the vital signs of patients with sepsis provides valuable insights to clinicians for timely interventions, such as administering stabilizing drugs or optimizing infusion strategies. Our research examines N-BEATS, an interpretable deep-learning forecasting model that can forecast 3 hours of vital signs for sepsis patients in intensive care units (ICUs). In this work, we use the N-BEATS interpretable configuration to forecast the vital sign trends and compare them with the actual trend to understand better the patient's changing condition and the effects of infused drugs on their vital signs. We evaluate our approach using the publicly available eICU Collaborative Research Database dataset and rigorously evaluate the vital sign forecasts using out-of-sample evaluation criteria. We present the performance of our model using error metrics, including mean squared error (MSE), mean average percentage error (MAPE), and dynamic time warping (DTW), where the best scores achieved are 18.52e-4, 7.60, and 17.63e-3, respectively. We analyze the samples where the forecasted trend does not match the actual trend and study the impact of infused drugs on changing the actual vital signs compared to the forecasted trend. Additionally, we examined the mortality rates of patients where the actual trend and the forecasted trend did not match. We observed that the mortality rate was higher (92\%) when the actual and forecasted trends closely matched, compared to when they were not similar (84\%).
\end{abstract}
%
%
\section{Introduction} 
Septic shock, the most severe form of sepsis, is characterized by profound circulatory and cellular abnormalities and is associated with a high mortality rate \cite{rhodes2017surviving,singer2016third}. Early detection and treatment of sepsis and septic shock are critical for improving patient outcomes, as delays in intervention can lead to a rapid decline in a patient's condition \cite{kumar2006duration}. While various methods have shown promise in predictive accuracy \cite{desautels2016prediction,komorowski2018artificial}, their interpretability remains a significant concern. Understanding and explaining a model's underlying mechanisms is crucial for gaining trust \cite{vilone2021quantitative,vilone2020comparative,rizzo2020empirical}, facilitating model debugging, and informing clinical decision-making \cite{arrieta2020explainable,vilone2021notions,vilone2022novel}. Interpretability and explainability methods for time series forecasting models can provide valuable insights into the relationships between vital signs and the risk of septic shock \cite{ribeiro2016should,rizzo2018comparative,vaswani2017attention,wachter2017counterfactual}. Several studies have been undertaken to incorporate model-agnostic explainability by leveraging rule-based \cite{vilone2021quantitative,vilone2020comparative} methods and argumentation \cite{longo2021examining,rizzo2018qualitative,rizzo2019inferential,rizzo2020empirical,rizzo2023comparing}. This paper explores the interpretability of vital sign forecasting models for patients with Sepsis and septic shock condition in critical care settings. Based on our knowledge, this work is one of the first to explore deep learning models to forecast the vital signs in the eICU dataset \cite{pollard2018eicu}. Further, we investigate the interpretability and explainability of patients' forecasted signals in conjunction with drug infusion. Our goal is to contribute to developing more interpretable and trustworthy models for septic shock prediction, ultimately improving patient outcomes.

\section{Methodology}
\subsection{Method}
In this paper, we utilize the N-BEATS \cite{oreshkin2020nbeats} interpretable architecture to forecast the seasonality and trend components. By accurately forecasting the trend component, we aim to provide valuable support to clinicians in monitoring the vital signs trend of patients and making informed decisions regarding medication administration. The interpretable configuration of N-BEATS consists of two distinct stacks that enable the direct incorporation of trend and seasonality decomposition within the model's architecture. This integration allows for a clearer interpretation of the stack outputs, as the model explicitly captures and represents the trend and seasonality components of the time series. As described in Equation \ref{eq:1} \cite{oreshkin2020nbeats}, the trend model is constrained to have a polynomial of small degree $p$, a function slowly varying across the forecast window. 

\begin{equation}\label{eq:1}
    \widehat{\mathbf{y}}=\sum_{i=0}^{p} \theta_{tr, i} t^{i}
\end{equation}

Here time vector $\mathbf{t}=[0,1,2, \ldots, H-2, H-1]^{T} / H$ is defined on a discrete grid running from 0 to $(H-1) / H$, forecasting $H$ steps ahead and $\theta_{tr}$ are polynomial coefficients predicted by a fully connected network. Similarly, the seasonality model, which captures the regular, cyclic and recurring fluctuation, is constrained with the Fourier basis as shown in Equation \ref{eq:2} \cite{oreshkin2020nbeats}, where $\theta_{sea}$ are Fourier coefficients predicted by a fully connected network.

\begin{equation}\label{eq:2}
   \widehat{\mathbf{y}}=\sum_{i=0}^{\lfloor H / 2-1\rfloor} \theta_{sea, i} \cos (2 \pi i t)+\theta_{sea, i+\lfloor H / 2\rfloor} \sin (2 \pi i t),
\end{equation}

\subsection{Dataset}
To evaluate our approach using N-BEATS, we conducted experiments on the publicly available eICU Collaborative Research Database \cite{pollard2018eicu}. The dataset includes information on 139,367 unique patients (53.96\% male, 45.95\% female, 0.09\% other/unknown). Our experiments focused on forecasting the mean blood pressure (MBP) of the patients diagnosed with sepsis or septic shock.
We perform data cleaning and preprocessing on our samples for the deep learning model. Missing values in the MBP are imputed using fill forward technique. We extract 9 hours of MBP before diagnosing sepsis or septic shock and split them into a 6-hour lookback horizon and a 3-hour forecasting horizon. Moving average smoothing with a window length of 3 is applied to remove noise. We remove samples with a standard deviation of  $\le$ 0.025, as they exhibit minimal variability. Additionally, we apply min-max scaling in the range of 0 to 190 to normalize the samples \cite{o2020characterizing}. After preprocessing, we obtain 4020 samples from 1442 patients.

\begin{table}
\caption{Performance of N-BEATS with Generic and Interpretable Configurations.} 
\label{tab1}
\centering

\begin{tabular}{|c|c|c|c|}
\hline
\textbf{Model Configuration} & \textbf{MSE ({1e-4})} & \textbf{MAPE}& \textbf{DTW ({1e-3})} \\
\hline
Presistence &  24.55 & 8.47 & 34.50 \\
N-BEATS -  Generic &  \textbf{18.52} & \textbf{7.60} & \textbf{17.63} \\    
N-BEATS -  Interpretable &  21.49 & 8.60 & 18.20 \\
\hline
\end{tabular}
\end{table}

\section{Results and Interpretability}
We assess the effectiveness of N-BEATS Generic and Interpretable configurations by analyzing their performance using three error metrics: Mean Squared Error (MSE), Mean Absolute Percentage Error (MAPE), and Dynamic Time Warping (DTW). To establish a baseline, we compare the forecasting outcomes with a simple persistence model. This model predicts future values of the time series by replicating the last observed value within the lookback horizon. The performance of the models on the test set are presented in Table \ref{tab1}.

\begin{figure}[t]
\centering
\includegraphics[height=.28\textwidth]{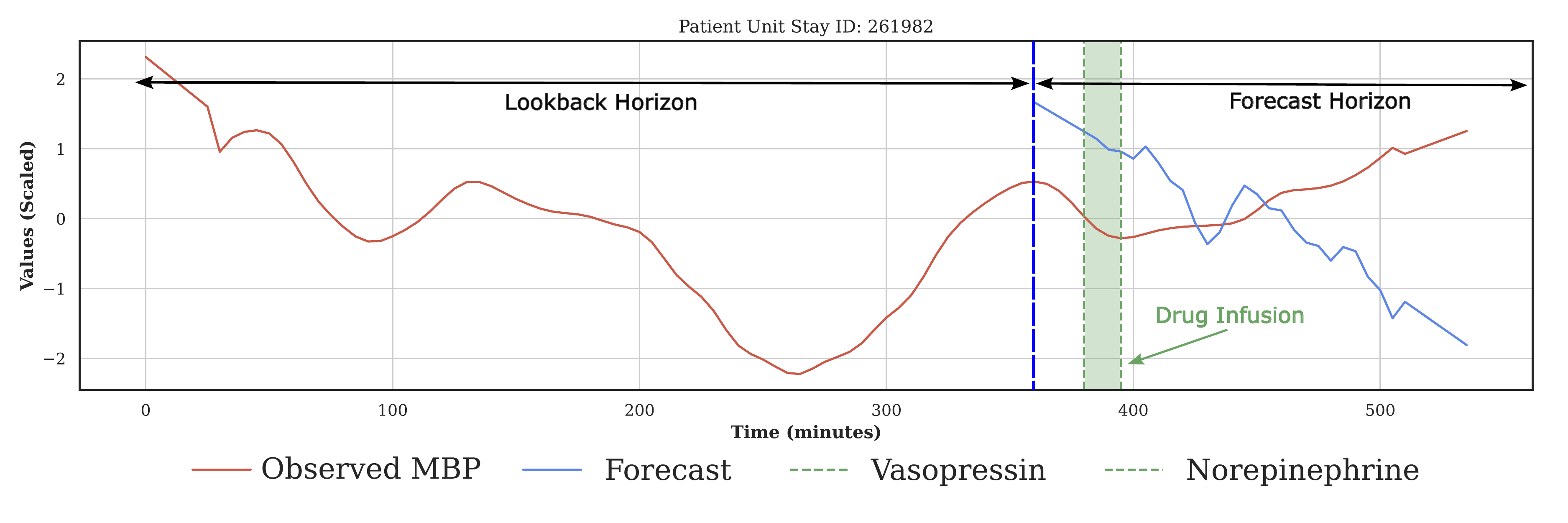}
\caption{The observed and forecasted trends of mean blood pressure (MBP) for a sample drawn from the subset with the highest DTW scores (top 25\%).} \label{fig1}
\end{figure}

Additionally, we investigate the extracted trend of the forecasts obtained using the N-BEATS interpretable configuration and analyze cases where the forecasted trend does not align with the actual trend. To better understand this discrepancy, we calculate the DTW score between the actual and the forecasted trends. We select the top 25\% of samples with the highest DTW scores, indicating a significant trend mismatch. We observe that, in several samples, a noticeable deviation between actual and forecasted trends occurs when drugs are administered after the training cut-off, as seen with patient ID 261982, who experienced an increased MBP trend due to the administration of vasoactive drugs like norepinephrine and vasopressin. Since the drug infusion information was an unobserved variable during the training of the forecasting model, the discrepancy in the actual and the forecasted trend can be attributed to the fact that the model is trained on historical vital sign data, which does not include the effects of drugs introduced after the training cut-off. Figure \ref{fig1} shows the observed MBP trend, the forecasted trend, the training cut-off, and the interval of drug infusion for a specific patient. Furthermore, our findings suggest that cases where the actual and forecasted trends matched had a higher mortality rate (92\%) compared to cases where the trends were dissimilar (84\%). Although not directly related, conducting further experiments could provide more insights into this aspect and explore its implications more extensively.

Exceptions to observed patterns can occur when drugs are administered before the specified cut-off, possibly due to drug-to-drug interactions. Conducting additional studies is crucial for understanding the causal inferences behind these interactions and better comprehending the relationship between medication administration, vital signs, and their subsequent effects.

\section{Conclusion}

In conclusion, our study utilized the eICU dataset to evaluate the forecasting performance of the interpretable N-BEATS model, highlighting the significance of accounting for drug infusion's influence on trends in ICU patients' vital signs. Future research will focus on developing approaches that integrate drug infusion information and investigate drug-to-drug interactions within the ICU context, aiming to enhance the overall performance of forecasting models in critical care. Addressing these challenges effectively can advance the understanding and application of forecasting methods, leading to improved patient care and better outcomes.

\bibliographystyle{splncs04}
%

\bibliography{reference}




\end{document}